\documentclass[10pt,twocolumn,letterpaper]{article}
\usepackage[accsupp]{axessibility}
\usepackage{iccv}
\usepackage{times}
\usepackage{microtype}
\usepackage{graphicx}
\usepackage{subfigure}
\usepackage{booktabs} 
\usepackage{multirow}
\usepackage{url}
\usepackage{amssymb}
\usepackage{xcolor}
\usepackage{pifont}
\usepackage{enumitem}
\usepackage{bm}
\usepackage{color}
\usepackage{amsmath}

\newcommand{\cmark}{\ding{51}}%
\newcommand{\xmark}{\ding{55}}%

\usepackage[pagebackref=true,breaklinks=true,letterpaper=true,colorlinks,bookmarks=false]{hyperref}

\iccvfinalcopy 

\def\iccvPaperID{9} 

\ificcvfinal\fi

\begin{document}

\title{Rethinking Content and Style: \\ Exploring Bias for Unsupervised Disentanglement}

\author{Xuanchi Ren$^{1}$\thanks{This work was done when Xuanchi Ren was an intern at MSRA} 
\quad  Tao Yang$^{2,3}$
\quad  Yuwang Wang$^3$\thanks{Corresponding author}  
\quad  Wenjun Zeng$^3$\\
$^1$HKUST  \qquad $^2$Xi'an Jiaotong University \qquad $^3$Microsoft Research Asia \\
}

\maketitle
\ificcvfinal\thispagestyle{empty}\fi

\begin{abstract}
Content and style (C-S) disentanglement intends to decompose the underlying explanatory factors of objects into two independent subspaces. From the unsupervised disentanglement perspective, we rethink content and style and propose a formulation for unsupervised C-S disentanglement based on our assumption that different factors are of different importance and popularity for image reconstruction, which serves as a data bias. The corresponding model inductive bias is introduced by our proposed \emph{C-S disentanglement Module} (\texttt{C-S DisMo}), which assigns different and independent roles to content and style when approximating the real data distributions. Specifically, each content embedding from the dataset, which encodes the most dominant factors for image reconstruction, is assumed to be sampled from a shared distribution across the dataset. The style embedding for a particular image, encoding the remaining factors, is used to customize the shared distribution through an affine transformation. The experiments on several popular datasets demonstrate that our method achieves the state-of-the-art unsupervised C-S disentanglement, which is comparable or even better than supervised methods. We verify the effectiveness of our method by downstream tasks: image-to-image translation and single-view 3D reconstruction.
Project page at {\url{https://github.com/xrenaa/CS-DisMo}}.
\end{abstract}

\section{Introduction}

The disentanglement task aims to recover the underlying explanatory factors of natural images into different dimensions of latent space, and provide an informative representation for downstream tasks like image translation~\cite{TransGaGa,kotovenko2019content}, domain adaptation~\cite{li2019cross} and geometric attributes extraction~\cite{XingHGZW19}, etc.

In this work, we focus on content and style (C-S) disentanglement, where content and style represent two independent latent subspaces.  Most of the previous C-S disentanglement works~\cite{DrNet,cycleVAE, MLVAE, Lord} rely on supervision.
For example, Gabbay and Hoshen~\cite{Lord} achieve disentanglement by forcing images from the {\it same group} to share a common embedding. It is not tractable, however, to collect such a dataset (\textit{e.g.} groups of paintings with each group depicting the same scene in different styles, or groups of portraits with each group depicting the same person with different poses). To our best knowledge, the only exception is Wu et al.~\cite{WuCLQL19} which, however, forces the content to encode pre-defined geometric structure limited by the expressive ability of 2D landmarks. 

Previous works define the content and style based on either the supervision or manually pre-defined attributes. There is no general definition of content and style for \textit{unsupervised} C-S disentanglement. In this work, we define content and style from the perspective of VAE-based unsupervised disentanglement works~\cite{higgins2016beta,burgess2018understanding,FactorVAE,chen2018isolating}.
These methods try to explain the images with the latent factors, of which each controls only one interpretable aspect of the images. However, extracting all disentangled factors is a very challenging task, and Locatello et al.~\cite{LocatelloBLRGSB19} prove that unsupervised disentanglement is fundamentally impossible without inductive bias on both the model and data. Furthermore, these methods have limited down-stream applications due to poor image generation quality on real-world datasets.

Inspired by the observation that the latent factors have different degrees of importance for image reconstruction~\cite{burgess2018understanding}, we assume the disentangled factors are of different importance when modeling the real data distributions. Instead of finding all the independent factors, which is challenging, we make the problem tractable by defining \emph{content} as a group of factors that are the most important ones for image reconstruction across the whole dataset, and defining \emph{style} as the remaining ones. 
Take the human face dataset CelebA~\cite{celeba} as an example, as pose is a more dominant factor than identity for image reconstruction across the face dataset, content encodes pose, and identity is encoded by style. We further assume that each content embedding of the dataset is sampled from a shared distribution, which characterizes the intrinsic property of content. For the real-world dataset CelebA, the shared distribution of content (pose) is approximately a Standard Normal Distribution, where zero-valued embedding stands for the canonical pose.  For the synthetic dataset Chairs~\cite{chairs}, as each image is synthesized from equally distributed surround views, the shared distribution of content (pose) is approximately an Uniform Distribution. 

Based on the above definitions and assumptions, we propose a problem formulation for unsupervised C-S disentanglement, and a C-S \textbf{Dis}entanglement \textbf{Mo}dule (\textbf{\texttt{C-S DisMo}}) which assigns different and independent roles to content and style when approximating the real data distributions. Specifically, \texttt{C-S DisMo} forces the content embeddings of individual images to follow a common distribution, and the style embeddings are used to scale and shift the common distribution to match the target image distribution via a generator. With the above assumptions as the data inductive bias, and \texttt{C-S DisMo} as the corresponding model inductive bias, we achieve unsupervised C-S disentanglement with good image generation quality. Furthermore, we demonstrate the effectiveness of our disentangled C-S
representations on two down-stream applications, \textit{i.e.}, image-to-image translation and single-view 3D reconstruction.

\begin{figure*}[t]
\centering
\includegraphics[width=0.9\linewidth]{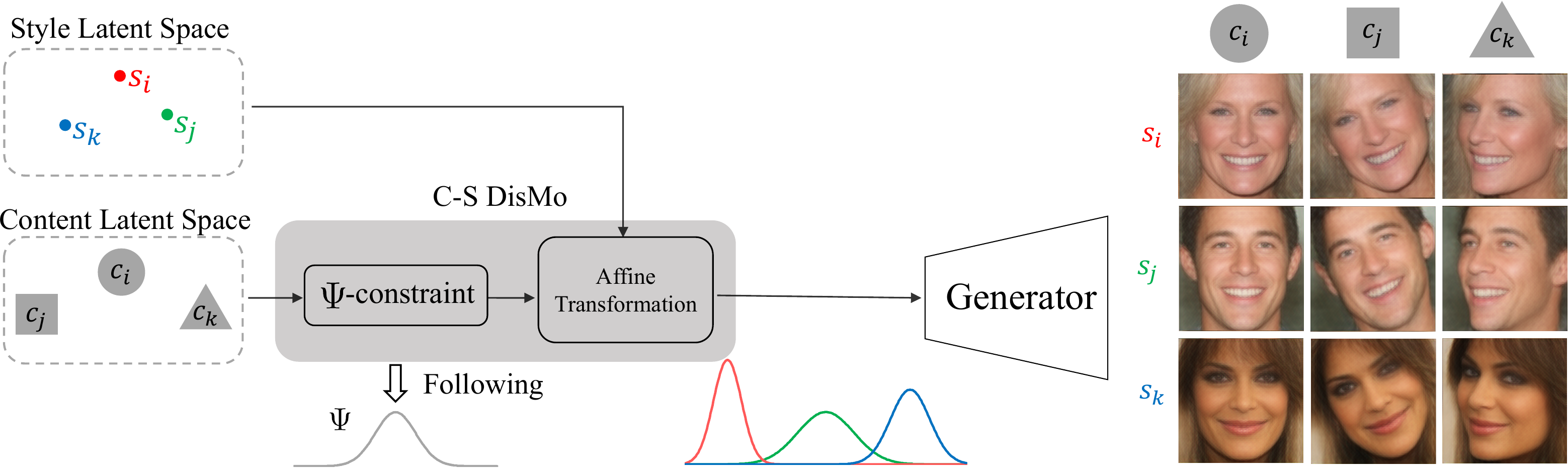}
\caption{
Overview of our method. Content embeddings $c_i,c_j,c_k$ are labelled with different shapes, and style embeddings $s_i,s_j,s_k$ are labelled with different colors. 
A C-S Disentanglement Module (\texttt{C-S DisMo}) is composed of a \emph{\textbf{$\Psi$-constraint }}and an \emph{\textbf{affine transformation}}. 
The \emph{\textbf{$\Psi$-constraint}} forces content embeddings to follow a shared distribution $\Psi$ and the \emph{\textbf{affine transformation}} scales and shifts the shared content
distribution with different styles (colors) as the Generator's input to approximate the target image distributions. 
Each image from $3\times3$ grids (right side) is generated with the content embedding from the column and style embedding from the row.}
\label{fig:model}
\vspace{-1em}
\end{figure*}

We follow Gabbay and Hoshen~\cite{Lord} to apply latent optimization to optimize the embeddings and the parameters of the generator. Please note that we only use the image reconstruction loss as the supervision; no human annotation is needed. We also propose to use instance discrimination as an auxiliary constraint to assist the disentanglement.  

The experiments on several popular datasets demonstrate that our method achieves the state-of-the-art (SOTA) unsupervised C-S disentanglement, which is comparable or even better than supervised methods. Furthermore, by simplifying the factors disentanglement problem into the C-S disentanglement problem, we achieve much better performance than the SOTA VAE-based unsupervised disentanglement method when modified for C-S disentanglement by manually splitting the factors into content and style.

\textbf{Main contributions.} The main contributions of our work are as follows: $(i)$ By rethinking content and style from the perspective of VAE-based unsupervised disentanglement, we achieve unsupervised C-S disentanglement by introducing both data and model inductive bias. $(ii)$ We propose the \texttt{C-S DisMo} to assign different and independent roles to content and style when modeling the real data distributions, and we provide several solutions for the distribution constraint of content. $(iii)$ We verify the effectiveness of our method by presenting two down-stream applications based on the well-disentangled content and style.

\section{Related Work}
\textbf{Unsupervised Disentanglement.} 
There have been a lot of studies on unsupervised disentangled representation learning~\cite{higgins2016beta,burgess2018understanding,FactorVAE,chen2018isolating}. These models learn disentangled factors by factorizing aggregated posterior. 
However, Locatello et al.~\cite{LocatelloBLRGSB19} prove that unsupervised disentanglement is impossible without introducing inductive bias on both the models and data. Therefore, these models are currently unable to obtain a promising disentangled representation. Inspired by these previous unsupervised disentanglement works, we revisit and formulate the unsupervised C-S disentanglement problem. We simplify the challenging task of extracting individual disentanglement factors into a manageable task: disentangling two groups of factors (content and style). 

\textbf{C-S Disentanglement.}
Originated from style transfer, most of the prior works on C-S disentanglement divide latent variables into two spaces relying on group supervision. To achieve disentanglement,  Mathieu et al.~\cite{DMathieuZZRSL16} and Szabo et al.~\cite{szabo_challenge} combine the adversarial constraint and auto-encoders. Meanwhile, VAE~\cite{VAE} is combined with non-adversarial constraints, such as cycle consistency~\cite{cycleVAE} and evidence accumulation~\cite{MLVAE}. Furthermore, latent optimization is shown to be superior to amortized inference for C-S disentanglement~\cite{Lord}. The only exception is Wu et al.~\cite{WuCLQL19}, which proposes a variational U-Net with structure learning for disentanglement in an unsupervised manner, but is limited by the expressive ability of 2D landmarks. In our work, we focus on the unsupervised C-S disentanglement problem and explore inductive bias for unsupervised disentanglement. 


\textbf{Image Translation.}
Image translation~\cite{MUNIT,FUNIT} between domains tries to decompose the latent space into domain-shared representations and domain-specific representations with the domain label of each image as supervision. 
The decomposition relies on the “swapping” operation and pixel-level adversarial loss without semantic level disentanglement ability. 
This pipeline fails in the unsupervised C-S disentanglement task on the single domain dataset due to lack of domain supervision. 
Our unsupervised C-S disentanglement task is to disentangle the latent space into content (containing most dominant factors typically carrying high-level semantic attributes) and style (containing the rest of the factors). 
We achieve disentangled content and style by assigning different roles to them without relying on domain supervision or the “swapping” operation.
We formulate the problem for a single domain but it can be extended to cross-domain to achieve domain translation without domain supervision, as shown in Figure~\ref{fig:edge}.

\section{Exploring Inductive Bias for Unsupervised C-S Disentanglement}
\subsection{Problem Formulation}
\label{sec:formulation}

For a given dataset $\mathcal{D}=\{I_i\}^N_{i=1}$, where $N$ is the total number of images, we assume each image $I_i$ is sampled from a distribution $P(\bm{x}|\bm{f}_1, \bm{f}_2, ..., \bm{f}_k)$, where $\{\bm{f}_i\}_{i=1}^k$ are the disentangled factors. Disentangling all these factors unsupervisedly is a challenging task, which has been proved to be fundamentally impossible without introducing the model and data inductive bias~\cite{LocatelloBLRGSB19}. Based on the observation that the factors play roles of different importance for image reconstruction~\cite{burgess2018understanding}, we assume $\{\bm{f}_i\}_{i=1}^k$ are of different importance and popularity for modeling the image distribution $P$.
We define the \emph{content} $\bm{c}$ as representing the most important factors across the whole dataset for image reconstruction and \emph{style} $\bm{s}$ as representing the rest ones. We assume c follows a shared distribution across the whole dataset, and assign each image $I_i$ a style embedding $s_i$ which parameterizes $P$ to be an image-specific distribution $P_{s_i}(\bm{x}|\bm{c})$. This serves as the data bias for our unsupervised C-S disentanglement.

With a generator $G_{\theta}$ that maps content and style embeddings to images, where $\theta$ is the parameter of the generator,  we further parameterize the target image distributions as $\{\hat{P}_{\theta,s_i}(\bm{x}|\bm{c})\}_{i=1}^N$. For each image $I_i$, we assign $c_i$ as the content embedding. All the content embeddings $\{c_i\}_{i=1}^N$ should conform the assumed distribution of content $\bm{c}$, which is denoted as $\Psi$.
Then we are able to estimate the likelihood of $I_i$ by $\hat{P}_{\theta,s_i}(\bm{x}|\bm{c}=c_i)$. 
Given the dataset D, our goal is to minimize the negative log-likelihood of $\hat{P}$:
\begin{equation}
\begin{aligned}
    & \min\limits_{\theta,c_i,s_i}-\sum_{i=1}^{N} \log{\hat{P}_{\theta,s_i}(\bm{x}=I_i|\bm{c} = c_i)}-\sum_{i=1}^{N} \log{\Psi(\bm{c}=c_i)}.
\end{aligned}
\label{equ:logp}
\end{equation}

\subsection{Proposed C-S Disentanglement Module}
\label{sec:architecture}
Here we propose a framework to address the formulated problem in Section~\ref{sec:formulation}.  We design a C-S \textbf{Dis}entanglement \textbf{Mo}dule (\textbf{\texttt{C-S DisMo}})  to assign different roles to content and style in modeling real data distributions according to their definitions (data bias) in Section 3.1, which servers as the corresponding model bias.

More specifically, 
as shown in Figure~\ref{fig:model}, a \texttt{C-S DisMo} is composed of a \emph{\textbf{$\Psi$-constraint }} to enforce content embeddings to conform to $\Psi$, which corresponds to the second term in Eq.~\ref{equ:logp}, and an \emph{\textbf{affine transformation}} serving to customize the shared content distribution into image-specific distributions. This module is followed by the generator to generate the target image.

The \emph{\textbf{affine transformation}} is inspired by the observation that the mean and variance of features carry individual information~\cite{GatysEB16,LiW16,LiWLH17,huang2017adain}. We use the style embeddings to provide the statistics to scale and shift content embedings as 
\begin{equation}
    z_i = f_{\sigma}(s_i)\cdot c_i + f_{\mu}(s_i),
    \label{equ:zi}
\end{equation}
where $f_{\sigma}$ and $f_{\mu}$ are two fully connected layers predicting the scalars for scaling and shifting respectively.
When $\hat{P}_{\theta,s_i}(\bm{x}|\bm{c}=c_i)$ is a Normal Distribution, Eq.~\ref{equ:logp} is equivalent to minimizing:
\begin{equation}
\label{equ:loss}
    \mathcal{L}_{CS} = {\sum_{i=1}^{N} \left \| I_i - G_\theta(z_i)\right \|} -\sum_{i=1}^{N} \log{\Psi(\bm{c}=c_i)},
\end{equation}
with the proof provided in the supplementary material. 

For the reconstruction term in Eq.~\ref{equ:loss}, we adopt a VGG perceptual loss~\cite{SimonyanZ14a, perceptual_loss}, which is widely used in unsupervised disentanglement methods~\cite{Wu_2020_CVPR, WuCLQL19}. 

\begin{figure*}[t]
\centering
\includegraphics[width=\linewidth]{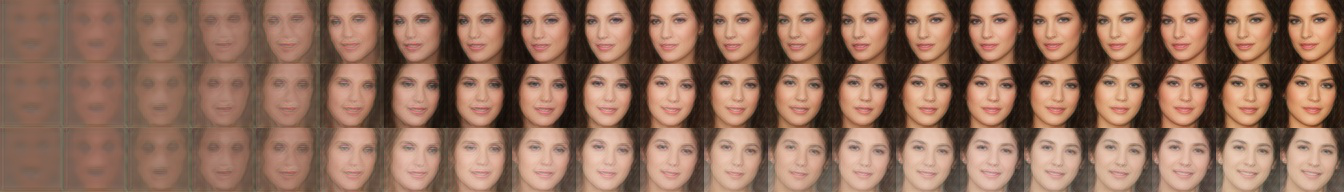}
\caption{Generated images at different training steps. The first and second rows share the same style embedding. The second and third rows share the same content embedding.}\label{fig:visual_training}
\vspace{-1.5em}
\end{figure*}

For the \emph{\textbf{$\Psi$-constraint}}, \textit{i.e.} the second term in Eq.~\ref{equ:loss}, we propose and study discrimination-based, NLL-based and normalization-based solutions. 
Choosing the form of $\Psi$, which can better approximate the ground truth content distribution of the dataset, can result in better disentanglement.
For real-world datasets, content is affected by a large number of random variables, we could assume the distribution of the content is nearly standard Normal Distribution.
We describe details of these solutions and related limitations according to the form of $\Psi$ below. 

\textbf{Discrimination-based solution} can be adopted when $\Psi$ has a tractable form for sampling. Inspired by adversarial learning~\cite{StyleGAN}, we propose to use a discriminator to distinguish between content embeddings $\{c_i\}_{i=1}^N$ (false samples) and items $\{\hat{c}_i\}_{i=1}^N$ sampled from $\Psi$ (true samples). 
When it is difficult for the discriminator to distinguish true from false, the content embeddings are likely to follow $\Psi$. 

\textbf{NLL-based solution} is inspired by flow-based generative models~\cite{glow}, and can be adopted when $\Psi = \mathcal{N}(\mu, \sigma^{2})$.
We can use negative log-likelihood (NLL) to optimize $\{c_i\}_{i=1}^N$ to follow $\Psi$ as
\begin{equation}
\min_{c_i} \frac{1}{N} \sum_{i = 1}^{N} \left(-\frac{\log2\pi}{2} - \log\sigma - \frac{(c_i - \mu)^2}{2\exp(2\log\sigma)} \right).
\end{equation}

\textbf{Normalization-based solution} can be adopted when $\Psi$ has one of the following specific forms: $i$) a Standard Normal Distribution $\mathcal{N}(0,I)$, and $ii$) a Uniform Distribution. To approximately follow the $\mathcal{N}(0,I)$ constraint, Instance Normalization (IN) is used to force the mean and variance of $c_i$ to be zeros and $I$ respectively. 
When $\Psi$ is a Uniform Distribution, we can use $\mathcal{L}_2$ normalization to force $\{c_i\}_{i=1}^N$ to follow Uniform Distribution approximately~\cite{muller1959note}. 

For these solutions, we show the qualitative and quantitative comparisons in Figure~\ref{fig:ablation_model} and Table~\ref{tbl:Ablation_tech} respectively to verify their effectiveness. 
Furthermore, discrimination-based and negative log-likelihood (NLL)-based solutions need extra optimization terms which introduce instability. In our work, we mainly adopt normalization-based solution to meet the \emph{\textbf{$\Psi$-constraint}}.

As shown in Figure~\ref{fig:model}, we can use the \texttt{C-S DisMo} before the generator, denoted as the \texttt{Single C-S DisMo framework}. 
We can also insert it before each layer of the generator to provide multiple paths for disentanglement, denoted as the \texttt{Multiple C-S DisMo framework}. 

\subsection{Demystifying C-S Disentanglement}
\label{sec:cs-ablation}
In this section, we perform some experiments to verify that the C-S disentanglement is achieved by introducing inductive bias on model (\texttt{C-S DisMo}) and data (our assumptions of the dataset). The experimental setting can be found in Section~\ref{sec:exp}.

To understand how \texttt{C-S DisMo} achieves disentanglement, we visualize the generated images during the training process of CelebA in Figure~\ref{fig:visual_training}. As the generated images show, a mean shape of faces is first learned. Then the faces start to rotate, which indicates the pose, as a dominant factor for generation, is disentangled as content. After that, the identity features emerge since the identity is disentangled as style for better image generation. 

\begin{figure}[t]
\centering
    \begin{tabular}{c@{\hspace{0.3em}}c}
{\includegraphics[width=0.48\linewidth]{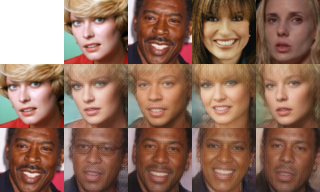}} &
{\includegraphics[width=0.48\linewidth]{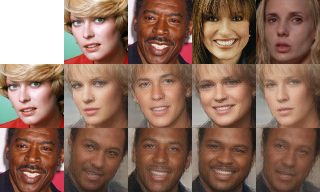}} \\
(a) Concatenation & (b) w/o $\Psi$-constraint \\
{\includegraphics[width=0.48\linewidth]{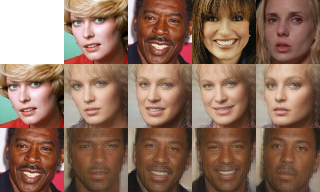}}&
{\includegraphics[width=0.48\linewidth]{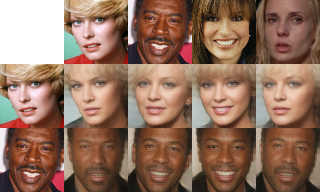}} \\
(c) Discrimination & (d) NLL \\
{\includegraphics[width=0.48\linewidth]{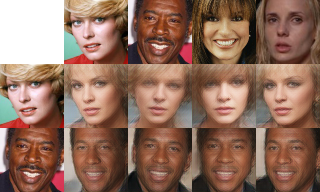}}&
{\includegraphics[width=0.48\linewidth]{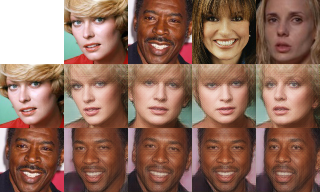}}
\\
(e) IN & (f) Multiple w/ IN \\
\end{tabular}
\caption{Ablation study of \texttt{C-S DisMo}. For each image, the content embedding is from the topmost image in the same column, and style embedding is from the leftmost image in the same row. A good disentanglement is that: horizontally, the style (identity) of the images is well maintained when the content (pose) varies, and vertically, the content of the images is well aligned when the style varies.}
    \label{fig:ablation_model}
\end{figure}

If we treat content and style equally, \textit{i.e.}, concatenating content and style embedding as the input of the generator, the network can hardly disentangle any meaningful information for the CelebA dataset, as shown in Figure~\ref{fig:ablation_model} (a). Our \texttt{Single C-S DisMo framework} with different solutions to meet \emph{\textbf{$\Psi$-constraint}} can disentangle the content (pose) and style (identity) of human faces, as shown in Figure~\ref{fig:ablation_model} (c)-(e). 
When \emph{\textbf{ $\Psi$-constraint }} is removed from \texttt{C-S DisMo}, the result is shown in Figure~\ref{fig:ablation_model} (b), where the pose and identity can not be disentangled. For the \texttt{Multiple C-S DisMo framework}, as multiple paths are provided, and the network has more flexibility to approximate the target image distribution, it outperforms the \texttt{Single C-S DisMo framework}, as shown in Figure~\ref{fig:ablation_model} (f).

\begin{figure}[t]
\centering
    \begin{tabular}{c@{\hspace{0.3em}}c}
{\includegraphics[width=0.48\linewidth]{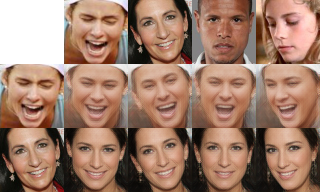}} &
{\includegraphics[width=0.48\linewidth]{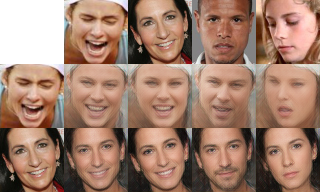}}\\
(a) IN & (b)  $\mathcal{L}_2$ Normalization \\
{\includegraphics[width=0.48\linewidth]{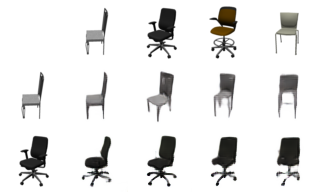}} &
{\includegraphics[width=0.48\linewidth]{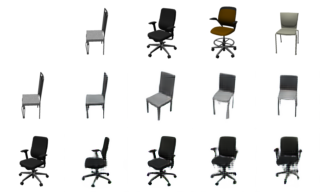}}
\\
(c) IN  & (d)  $\mathcal{L}_2$ Normalization
\end{tabular} 
\caption{Comparison of the disentanglement with different normalizations. Instance Normalization (IN) achieves better results on CelebA, \textit{e.g.}, the face identities are more alike.  $\mathcal{L}_2$ normalization outperforms on Chairs, where the shapes of chairs are more consistent in each row.}
\label{fig:ablation_model_dis}
\vspace{-1em}
\end{figure}

We conduct experiments to demonstrate that better disentanglement can be achieved by choosing a better form for $\Psi$. For the real-world dataset CelebA, the distribution of pose is better modeled as a Standard Normal Distribution. As Figure~\ref{fig:ablation_model_dis} (a) and (b) show, IN achieves better disentanglement than  $\mathcal{L}_2$.
For the synthetic Chairs~\cite{chairs} dataset, the distribution of pose is close to Uniform Distribution rather than Standard Normal Distribution. Figure~\ref{fig:ablation_model_dis} (c) and (d) show that the  $\mathcal{L}_2$ normalization results in better consistency of identity and pose. 

\subsection{Auxiliary Loss Function}
\label{sec:aux-loss}
In addition to the $\mathcal{L}_{CS}$ in Eq.~\ref{equ:loss}, we propose two auxiliary loss functions to help the model to better disentangle C-S.

\textbf{Instance discrimination.}
Instance discrimination can discover image-specific features~\cite{wu2018unsupervised_instancediscrimination}. The image-specific feature corresponds to style according to our definition. 
Inspired by this, we first pretrain a ResNet-18~\cite{heZRS16} $\Phi$ unsupervisedly with the method in \cite{wu2018unsupervised_instancediscrimination} and define a collection of layers of $\Phi$ as $\{ \Phi_{l} \}$.
Given two images $I_i$ and $I_j$, we mix the embeddings to generate $u = G_\theta(s_i, c_j)$ and $v = G_\theta(s_j, c_i)$. For samples sharing the same style embedding, we enforce the feature distance in $\Phi$ between them to be close. This loss term can be written as
\begin{equation}
    \mathcal{L_{ID}} = \sum_{l}\lambda_{l}(\Arrowvert\Phi_{l}(u)-\Phi_{l}(x)\Arrowvert_{1} + \Arrowvert \Phi_{l}(v)-\Phi_{l}(y)\Arrowvert_{1}),
\end{equation}
where $x = G_\theta(s_i, c_i)$ and $y = G_\theta(s_j, c_j)$. The hyperparameters $\{\lambda_{l}\}$ balance the contribution of each layer $l$ to the loss. $\{\lambda_l\}$ are set to be $[1,1,1,1,1]$.

\textbf{Information bottleneck.}
Burgess et al.~\cite{burgess2018understanding} propose improving the disentanglement by controlling the capacity increment. This motivated us to control the information bottleneck capacity of content and style to help to avoid leakage. We introduce an information bottleneck given by
\begin{equation}
    \mathcal{L_{IB}} =  \gamma_s \Arrowvert s^2 - C_s \Arrowvert_{1} + \gamma_c \Arrowvert c^2 - C_c \Arrowvert_{1}
\end{equation}
where $C_s$ and $C_c$ are the information capacity controlling the amount of information of the content and style, respectively. During training, $C_s$ and $C_c$ increase linearly. The rate of increase is controlled by the increase steps and the maximum value. By controlling the increase rate, the content is forced to encode information first so that the learning process is more consistent with our assumptions.

\begin{figure*}[t]
\centering
\begin{tabular}{c}
{\includegraphics[width=\linewidth]{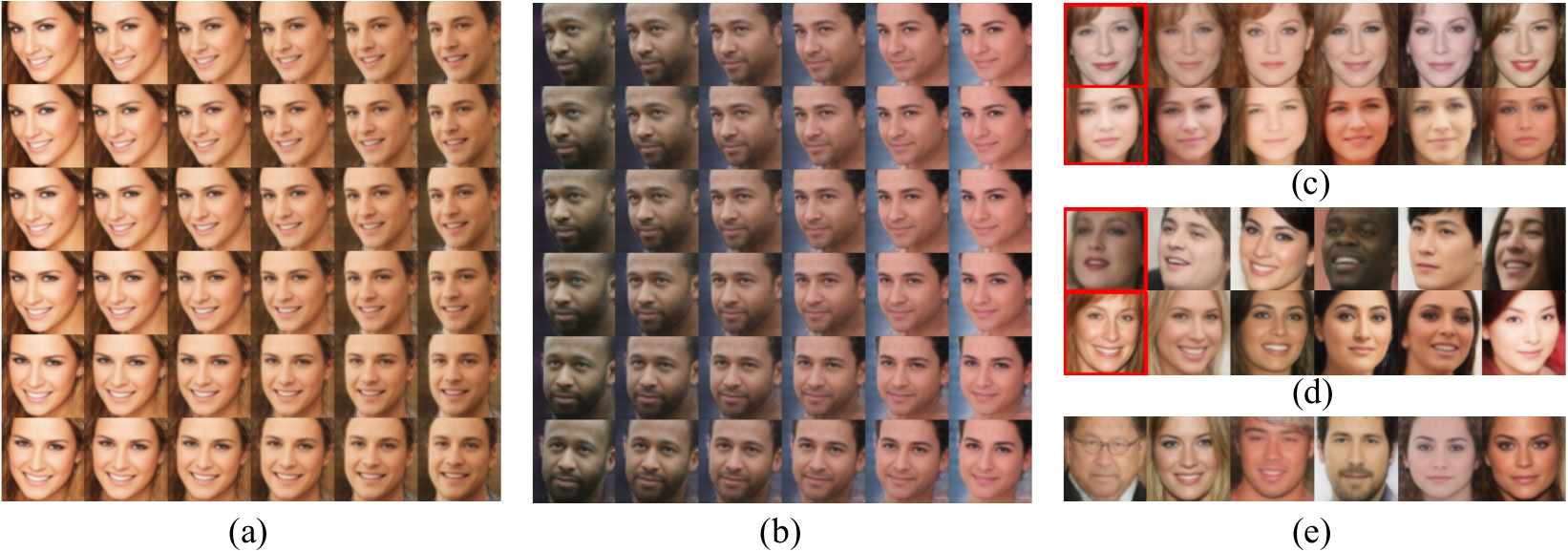}}
\end{tabular} 
\caption{
Demonstrations of the content and style space by interpolation (a \& b) and retrieval (c-e).}
\label{fig:interpolation}
\vspace{-1em}
\end{figure*}

\begin{figure}[t]
\centering
\renewcommand{\arraystretch}{0.8}
\begin{tabular}{c@{\hspace{0.3em}}c@{\hspace{0.3em}}c}
{\includegraphics[width=0.32\linewidth]{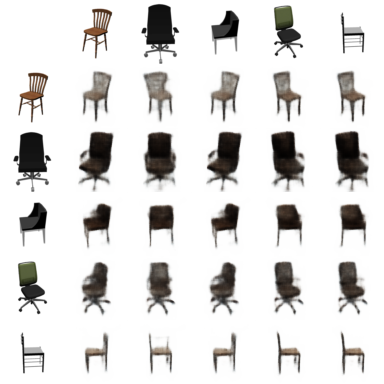}} &
{\includegraphics[width=0.32\linewidth]{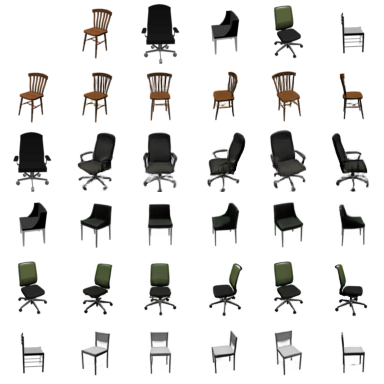}} &
{\includegraphics[width=0.32\linewidth]{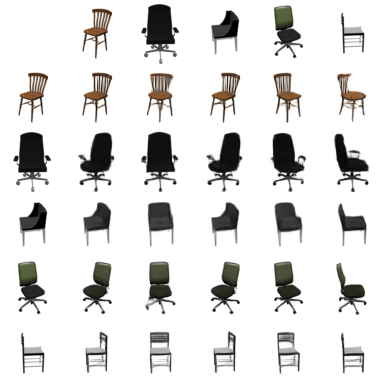}}\\
{\includegraphics[width=0.32\linewidth]{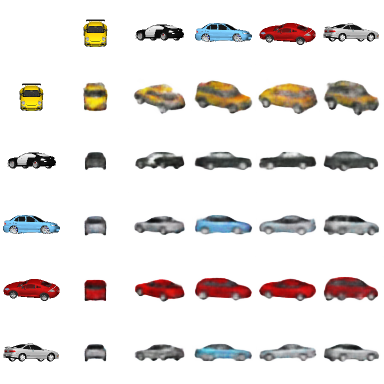}} &
{\includegraphics[width=0.32\linewidth]{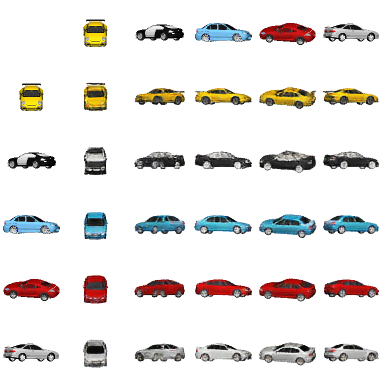}} &
{\includegraphics[width=0.32\linewidth]{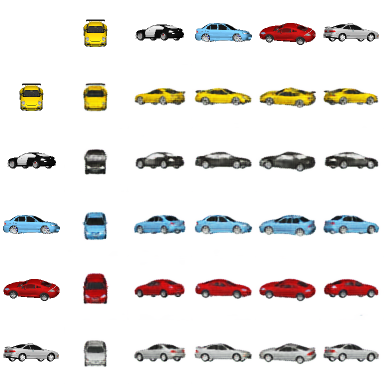}} \\
{\includegraphics[width=0.32\linewidth]{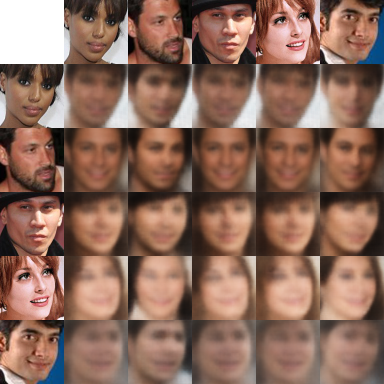}} &
{\includegraphics[width=0.32\linewidth]{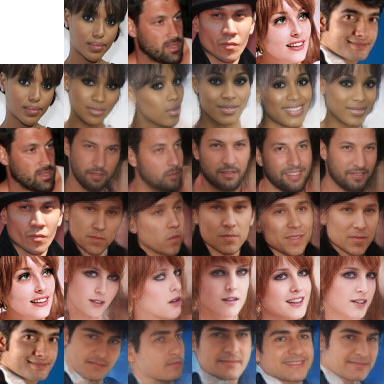}} &
{\includegraphics[width=0.32\linewidth]{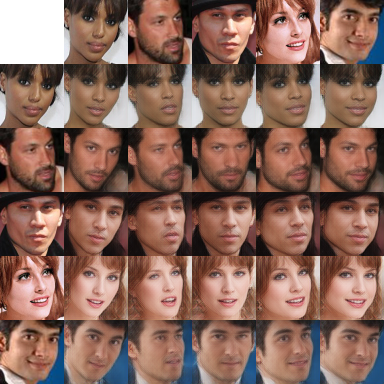}}\\
FactorVAE  & Lord  & Ours\\
\end{tabular} 
\caption{Comparison of visual analogy results on Chairs, Car3D and CelebA (from top to bottom). Zoom in for details.}
\label{fig:visual}
\vspace{-1 em}
\end{figure}

\textbf{Full objective.} Therefore, our full objective is
\begin{equation}
    w_{CS}\mathcal{L}_{CS}+w_{IB}\mathcal{L}_{IB}+ w_{ID}\mathcal{L}_{ID},
\end{equation}
where hyperparameters $w_{CS}$, $w_{IB}$, and $w_{ID}$ represent the weights for each loss term respectively. 

\section{Experiments}
\label{sec:exp}
In this section, we perform quantitative and qualitative experiments to evaluate our method.  
We test our method on several datasets: \textbf{Car3D}~\cite{car3d}, \textbf{Chairs}~\cite{chairs} and \textbf{CelebA}~\cite{celeba}. For these three datasets, pose is the most dominant factor and encoded by content.
For details of the datasets, please refer to the supplementary material.

\textbf{Baselines.}
We choose several SOTA group-supervised C-S disentanglement benchmarks for comparisons: 
Cycle-VAE~\cite{cycleVAE}, DrNet~\cite{DrNet} and Lord~\cite{Lord}.
We select the only unsupervised C-S disentangled method Wu et al.~\cite{WuCLQL19}~\footnote{There is no open-sourced implementation for it. We modify \url{https://github.com/CompVis/vunet} and provide pseudo ground truth landmarks to the network. Thus it becomes semi-supervised actually.}.
We choose one VAE-based unsupervised disentanglement method: FactorVAE~\cite{FactorVAE}. For FactorVAE, according to our definition of content and style, we manually traverse the latent space to select the factors related to pose as content and treat the other factors as style, for all these three datasets.
More details for baselines are presented in the supplementary material.

\subsection{Quantitative Experiments}
\label{sec:quanti}
We compare our method (\texttt{Multiple C-S DisMo framework}) with the baselines on {Car3D}, {Chairs} and {CelebA}. 

\textbf{Content Transfer Metric.}
To evaluate our method's disentanglement ability, we follow the protocol of Gabbay and Hoshen~\cite{Lord} to measure the quality of content transfer by LPIPS~\cite{LPIPS}. Details are presented in Appendix A. 
The results are shown in Table~\ref{tbl:Content}. We achieve the best performance among the unsupervised methods, even though pseudo labels are provided for Wu et al.\cite{WuCLQL19}. Our method significantly outperforms FactorVAE, which verifies the effectiveness of our formulation: simplifying the problem from disentangling factors to disentangling content and style.
Furthermore, our method is comparable to or even better than the supervised ones.

\textbf{Classification Metric.}
Classification accuracy is used to evaluate disentanglement in \cite{DrNet, cycleVAE, Lord}. 
we follow the protocol of Jha et al.~\cite{cycleVAE}.
Low classification accuracy indicates small leakage between content and style.
Without content annotations for CelebA, we regress the position of the facial landmarks from the style embeddings instead.
The results are summarized in Table~\ref{tbl:Classification}. Though without supervision, the performance of our method is still comparable to several other methods. We note that the classification metric may not be appropriate for disentanglement, which is also observed in Liu et al.\cite{liuS20}. The observation is that the classification metric is also influenced by information capacity and dimensions of embeddings. For FactorVAE, the poor reconstruction quality indicates that the content and style embeddings encode little information that can hardly be identified. The dimensions of the content and style embeddings of different methods vary from ten to hundreds, and a higher dimension usually leads to easier classification. 
 
\begin{table}[t]
\begin{center}
\resizebox{\linewidth}{!}{
\begin{tabular}{c@{\hspace{1em}}c@{\hspace{2em}}c@{\hspace{3em}}c@{\hspace{3em}}c}
\toprule
Method & Supervision & Cars3D & Chairs& CelebA\\
\midrule
DrNet~\cite{DrNet} & \multirow{3}{1em}{{\color{green} \cmark}}& 0.146 & 0.294 & 0.221 \\
Cycle-VAE~\cite{cycleVAE} & & 0.148& 0.240 & 0.202 \\
Lord~\cite{Lord} & & 0.089& \textbf{0.121} & 0.163 \\
\midrule
FactorVAE~\cite{FactorVAE}& \multirow{3}{1em}{{\color{red} \xmark}} &0.190 & 0.287 &0.369 \\
Wu et al.~\cite{WuCLQL19}$^1$ & & -- & -- & 0.185\\
Ours & &\textbf{0.082} & 0.190 &\textbf{0.161}\\
\bottomrule
\end{tabular}
}
\end{center}
\caption{Performance comparison in content tranfer metric (lower is better). For Wu et al.~\cite{WuCLQL19}$^1$, we provide pseudo facial landmarks, and there are no suitable landmarks for cars and chairs.}
\label{tbl:Content}
\vspace{-0.5em}
\end{table}

\begin{table}[t]
\begin{center}
\resizebox{\linewidth}{!}{
\begin{tabular}{c@{\hspace{2em}}c@{\hspace{2em}}c@{\hspace{1em}}c@{\hspace{1em}}c@{\hspace{1em}}c@{\hspace{1em}}c@{\hspace{1em}}c}
\toprule
\multirow{2}{4.5em}{~~Method} & \multirow{2}{4.5em}{Supervision} & \multicolumn{2}{c}{Cars3D} & \multicolumn{2}{c}{Chairs}& \multicolumn{2}{c}{CelebA}\\
\cmidrule(lr){3-8}
  & & $s \rightarrow c$ & $s \leftarrow c$& $s \rightarrow c$& $s \leftarrow c$& $R(s) \rightarrow c$& $s \leftarrow c$ \\
\midrule
DrNet~\cite{DrNet}& \multirow{3}{1em}{{\color{green} \cmark}} & 0.27& \textbf{0.03}& 0.06& 0.01& 4.99& \textbf{0.00} \\
Cycle-VAE~\cite{cycleVAE} & & {0.81} & {0.77} & {0.60}& {0.01}& 2.80 &{0.12}  \\
Lord~\cite{Lord} & & \textbf{0.03} & {0.09} & \textbf{0.02}& {0.01}& 4.42 &{0.01}  \\
\midrule
FactorVAE~\cite{FactorVAE} &\multirow{3}{1em}{{\color{red} \xmark}} & 0.07& 0.01 &0.14 &0.01 & 5.34& \textbf{0.00}\\
Wu et al.~\cite{WuCLQL19}$^1$& & --&-- &-- &-- & \textbf{5.42} &0.11\\
Ours & &0.33 &0.24 & 0.66& 0.05& 4.15 &0.05\\
\bottomrule
\end{tabular}
}
\end{center}
\caption{ Classification accuracy of  style labels from content codes ($s \leftarrow c$) and of content labels from style codes ($s \rightarrow c$) (lower is better). For Wu et al.~\cite{WuCLQL19}, we provide pseudo ground truth landmarks. Note that the column ($R(s) \rightarrow c$) presents the error of face landmark regression from the style embeddings (higher is better).}
\label{tbl:Classification}
\vspace{-0.5em}
\end{table}

\subsection{Qualitative Experiments}

\textbf{Disentanglement \& Alignment.} In Figure~\ref{fig:interpolation} (a) and (b), we conduct linear interpolation to show the variation in the two embedding spaces. Horizontally, with the interpolated style embeddings, the identity (style) is changed smoothly while the pose (content) is well maintained. Vertically, the identity remains the same as the pose changes. We have the following observations: $(i)$ The learned content and style spaces are continuous. $(ii)$ Columns of the left and right figures share the same pose, suggesting that the learned content spaces are well aligned. $(iii)$ Factors encoded by style is maintained when changing the content embeddings and vice versa, suggesting the good disentanglement.

We perform retrieval on the content and style latent spaces, respectively. 
As shown in Figure~\ref{fig:interpolation} (c) and (d), given a query image (labeled with a red box), its nearest neighbors in the content space share the same pose but have different identities, which reveals the content space is well aligned. To better identify the faces, we let the query's nearest neighbors in the style space share the same pose, and the generated faces look very similar, revealing that the style is well maintained. 
As shown in Figure~\ref{fig:interpolation} (e), zero-valued content embedding result in a canonical view.
As we assume that the pose distribution of faces is $\mathcal{N}(0, I)$, the canonical views are the most common pose in the dataset, and the zero-valued content embedding has the largest likelihood accordingly.

\begin{figure}[t]
\centering
\begin{tabular}{cc}
\rotatebox{90}{\hspace{4mm} FactorVAE}&
{\includegraphics[width=0.90\linewidth]{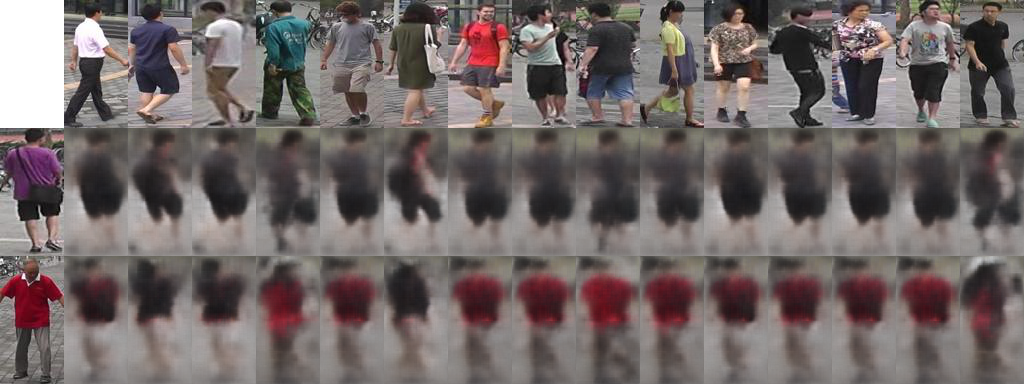}} \\
\rotatebox{90}{\hspace{4mm} Ours}&
{\includegraphics[width=0.90\linewidth]{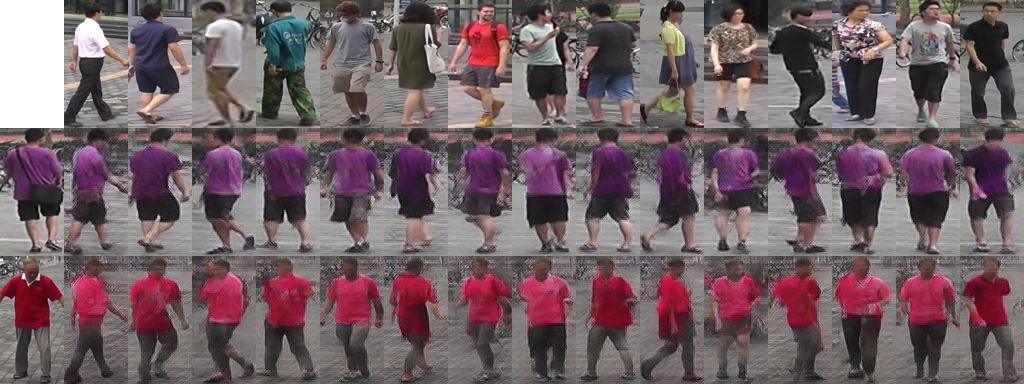}}
\end{tabular} 
\caption{Comparison of visual analogy results on Market-1501 dataset. 
Our method outperforms FactorVAE~\cite{FactorVAE} significantly.
}
\label{fig:reid_more}
\vspace{-1em}
\end{figure}

\begin{figure*}[t]
\centering
\begin{tabular}{c@{\hspace{1em}}c@{\hspace{0em}}c@{\hspace{0em}}c@{\hspace{1em}}c@{\hspace{1em}}c@{\hspace{1em}}c}
{\includegraphics[width=0.12\linewidth]{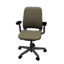}} &
{\includegraphics[width=0.12\linewidth]{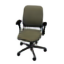}} &
{\includegraphics[width=0.12\linewidth]{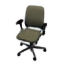}} &
{\includegraphics[width=0.12\linewidth]{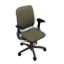}} &
{\includegraphics[width=0.12\linewidth]{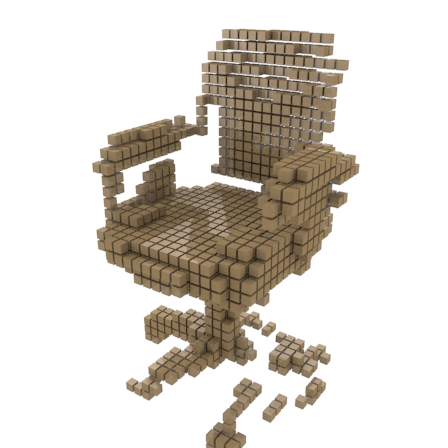}} &
{\includegraphics[width=0.12\linewidth]{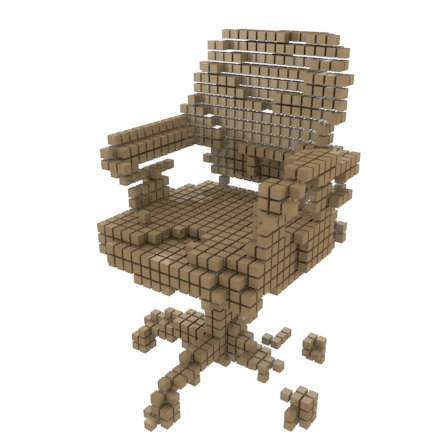}} &
{\includegraphics[width=0.12\linewidth]{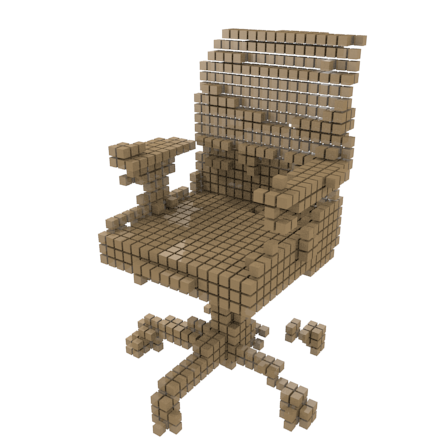}} \\
Input & \multicolumn{3}{c}{Our generated multi-view} & Single & Ours & GT
\end{tabular}
\caption{3D reconstruction results on Chairs. Single: the object reconstructed by only Input. Ours: the object reconstructed from multi-view inputs generated by our method from Input. GT: the object reconstructed by the ground truth of multi-view inputs. }
\label{fig:3d_chair}
\end{figure*}


\textbf{Visual Analogy \& Comparison.}
Visual analogy~\cite{car3d} is to switch style and content embeddings for each pair. We show the visual analogy results of our method against FactorVAE (typical unsupervised baseline) and Lord (strongest supervised baseline) in Figure~\ref{fig:visual} on Chairs, Car3D, and CelebA. The results show that FactorVAE on all datasets is of poor generation quality and bad content transfer.
On Cars3D, Lord's results have artifacts (\textit{e.g.}, third column), and its style embeddings could not encode the color information of the test images (\textit{e.g.}, fourth row). On CelebA, the transfer result of Lord is not consistent, \textit{e.g.}, the content embedding controls facial expression in the fifth column, while other content embeddings do not control expression. Our method achieves comparable pose transfer to Lord and maintains the identities of the images. 
Furthermore, we show our results on the Market-1501~\cite{zheng2015scalable} dataset in Figure~\ref{fig:reid_more}, which demonstrates our method can disentangle the human pose and the appearance even though the skeletons have large variances.
For more results (including on other datasets), please refer to the supplementary material.

\subsection{Ablation Study}
\label{sec:abl_study}

\textbf{Choice of $\Psi$-constraint.} Beside the qualitative experiment shown in Figure~\ref{fig:ablation_model_dis},
we perform ablation study on CelebA to evaluate different solutions for \emph{\textbf{$\Psi$-constraint}} introduced in Section~\ref{sec:architecture}. 
In this subsection, we do not use auxiliary loss functions.
As shown in Table~\ref{tbl:Ablation_tech}, all the solutions can achieve the SOTA performance in terms of content transfer metric, which means that the \emph{\textbf{$\Psi$-constraint}} for content embeddings is crucial. This result further verifies that our definition is reasonable. 
For the classification metric, the results of discrimination-based and NLL-based solutions are relatively poor due to the reasons discussed in Section~\ref{sec:quanti}. The normalization-based solution achieves the best results on all the metrics. We believe that is because the normalization-based solution does not introduce an extra optimization term, which may hurt the optimization process and limit the expressive ability of embeddings.  

\begin{figure*}[t]
\centering
\begin{tabular}{c@{\hspace{3mm}}c@{\hspace{3mm}}c}
{\includegraphics[width=0.32\linewidth]{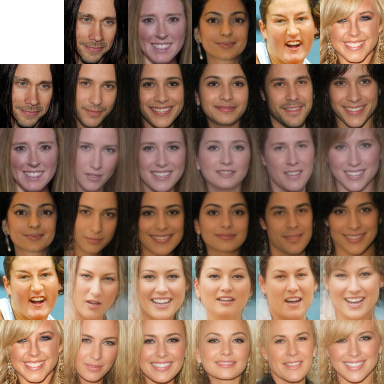}} &
{\includegraphics[width=0.32\linewidth]{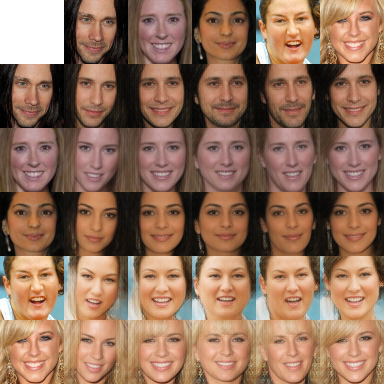}} &
{\includegraphics[width=0.32\linewidth]{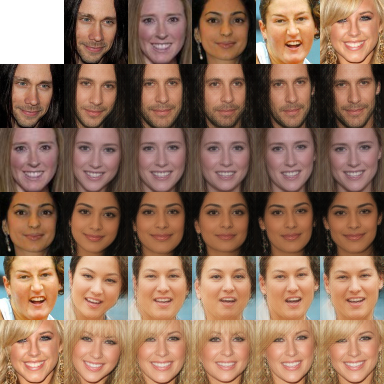}} \\
(a) $d_s = 128, d_c = 256$ & (b) $d_s = 256, d_c = 256$ & (c) $d_s = 256, d_c = 128$\\
\end{tabular} 
\caption{Study on the influence of size of embeddings. $d_s$ is the size of style embedding and $d_c$ is the size of content embedding.
}
\label{fig:leak}
\end{figure*}

\textbf{Choice of size of embeddings.} We also conduct experiments on the influence of the size of embeddings. We empirically set the size of style embedding $d_s$ to $256$ and the size of content embedding $d_c$ to $128$, which achieves good performance on all the datasets. Here, we demonstrate that we can also control the role of style by adjusting the size of the embeddings, as shown in Figure~\ref{fig:leak}.
For Figure~\ref{fig:leak} (a), the content embeddings contain the shape of the face, facial expression, and pose. 
For Figure~\ref{fig:leak} (b), the content embeddings contain the shape of the face and facial expression. 
For Figure~\ref{fig:leak} (c), which is the setting used in our paper, the content embeddings contain pose.


\subsection{Unseen Images Inference}
\begin{figure}[t]
\centering
\begin{tabular}{c@{\hspace{0.5em}}c}
{\includegraphics[width=0.48\linewidth]{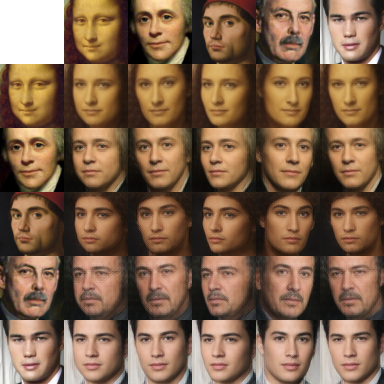}} &
{\includegraphics[width=0.48\linewidth]{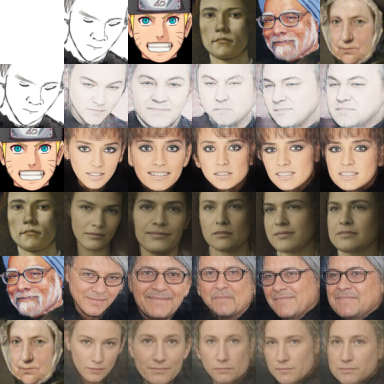}} 
\end{tabular}
\caption{Inference for unseen images. Our method performs well on images from different domains: painting and cartoon.}
\label{fig:unseen}
\vspace{-1em}
\end{figure}

Our method can be generalized to the held-out data.
A solution is to train two encoders to encode images to the content and style spaces respectively. We train a style encoder $E_s$ and a content encoder $E_c$ by minimizing
\begin{equation}
	\mathcal{L}_{E} = \sum_{i=1}^{N} \Arrowvert E_s(I_i) - s_i \Arrowvert_{1} + \Arrowvert E_c(I_i) - c_i  \Arrowvert_{1}.
\end{equation}
We apply our model trained on the CelebA dataset to faces collected by Wu et al.~\cite{Wu_2020_CVPR} including paintings and cartoon drawings. As shown in Figure~\ref{fig:unseen}, our method can be well generalized to unseen images from different domains.

\begin{figure}[t]
\centering
\begin{tabular}{c@{\hspace{1em}}c@{\hspace{1em}}c}
{\includegraphics[width=0.40\linewidth]{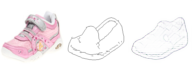}} & \quad \quad
{\includegraphics[width=0.40\linewidth]{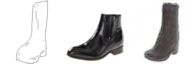}} \\


{\includegraphics[width=0.40\linewidth]{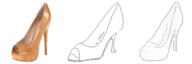}} & \quad \quad
{\includegraphics[width=0.40\linewidth]{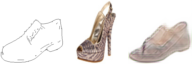}} \\
\end{tabular} 
\caption{Examples of translating shoes to edge (left column) and translating edges to shoes (right column). Triplet order (left to right) is: content, style, translation.}
\label{fig:edge}
\vspace{-1em}
\end{figure}

\begin{table}[t]
\begin{center}
\resizebox{\linewidth}{!}{
\begin{tabular}{c@{\hspace{3em}}c@{\hspace{2em}}cc}
\toprule
\multirow{2}{*}{Method} & \multirow{2}{*}{Content transfer metric $\downarrow$ } &  \multicolumn{2}{c}{Classification metric}\\
\cmidrule(lr){3-4}
 & & $R(s) \rightarrow c$ $\uparrow$ & $s \leftarrow c$  $\downarrow$  \\
\midrule
\multicolumn{1}{l}{Single} & 0.204 & 3.03 & 0.06 \\
\multicolumn{1}{l}{Single w/ Disc} & 0.178 & 2.97 & 0.14 \\
\multicolumn{1}{l}{Single w/ NLL} & 0.171 & 2.98 & 0.09\\
\multicolumn{1}{l}{Single w/ IN} & \textbf{0.166} & \textbf{3.46} & \textbf{0.04} \\
\bottomrule
\end{tabular}}
\end{center}
\caption{Ablation study for different solutions for \emph{\textbf{$\Psi$-constraint}} on Celeba. Disc means discrimination-based solution. 
}
\label{tbl:Ablation_tech}
\vspace{-3mm}
\end{table}

\subsection{Extension for Applications}
In this work, we explore two applications of C-S disentanglement. 
For 3D reconstruction, single-view settings lack reliable 3D constraints~\cite{WuBCLSLN019}. Base on our disentangled representations, we can generate multi-view from a single view.
On Chairs, we adopt Pix2Vox~\cite{xie2019pix2vox}, a framework for single-view, and multi-view 3D reconstruction to verify the effectiveness of our method.
As shown in Figure~\ref{fig:3d_chair}, the 3D objects reconstructed from multi-view generated from our method are much better than those reconstructed from a single view, and even comparable to those reconstructed from ground-truth multi-view.

For domain translation, our method can work on the images merged from two domains without using any domain label. As shown in Figure~\ref{fig:edge}, based on the disentangled content (edge structure) and style (texture), we can translate edge images into shoe images and vice versa. 

\section{Conclusion}
We propose a definition for content and style and a problem formulation for unsupervised C-S disentanglement. Based on the formulation, \texttt{C-S DisMo} is proposed to assign different and independent roles to content and style when approximating the real data distributions. 
Our method outperforms other unsupervised approaches and achieves comparable or even better performance than the SOTA supervised methods. As for the limitation, we fail on datasets containing multiple categories with large appearance variation, \textit{e.g.}, CIFAR-10~\cite{krizhevsky2009learning}, which does not match our assumption. As for application, our method could be adopted to help downstream tasks, \textit{e.g.}, domain translation, single-view 3D reconstruction, etc.

{\small
\bibliographystyle{ieee_fullname}
\bibliography{egbib}
}

\end{document}